\documentclass[preprints,article,accept,pdftex,moreauthors,captions=figureheading]{Definitions/mdpi} 
\firstpage{1} 
\pubvolume{1}
\issuenum{1}
\articlenumber{0}
\pubyear{2022}
\copyrightyear{2022}
\datereceived{} 
\dateaccepted{} 
\datepublished{} 
\hreflink{https://doi.org/} % If needed use \linebreak
\Title{Interactive Visualization and Representation Analysis Applied to Glacier Segmentation}

\TitleCitation{Interactive Visualization and Representation Analysis Applied to Glacier Segmentation}

\Author{Minxing Zheng $^{1}$%\orcidA{}
, Xinran Miao $^{1}$ and Kris Sankaran $^{1,}$*}

\AuthorNames{Minxing Zheng, Xinran Miao and Kris Sankaran}

\AuthorCitation{Zheng, M.; Miao, X.; Sankaran, K.}
\address{%
$^{1}$ \quad Department of Statistics, University of Wisconsin Madison, Madison, WI 53706, USA; mzheng54@wisc.edu (M.Z.); xmiao27@wisc.edu (X.M.)}

\corres{Correspondence: ksankaran@wisc.edu}

\abstract{Interpretability has attracted increasing attention in earth observation problems. We apply interactive visualization and representation analysis to guide interpretation of glacier segmentation models. We visualize the activations from a U-Net to understand and evaluate the model performance. We build an online interface using the Shiny R package to provide comprehensive error analysis of the predictions. Users can interact with the panels and discover model failure modes. Further, we discuss how visualization can provide sanity checks during data preprocessing and model training.}


\begin{document}
%%%%%%%%%%%%%%%%%%%%%%%%%%%%%%%%%%%%%%%%%%
%\setcounter{section}{-1} %% Remove this when starting to work on the template.
\section{Introduction}

Advances in machine learning for remote sensing have enabled automatic tracking of changes of glacier cover. Researchers have proposed to implement deep learning models on satellite images to identify the positions and boundaries of glaciers. These approaches have been shown to have improved performance relative to non-deep learning baselines.
However, the literature on interpretability of these black-box models is limited. This gap is problematic, because interpretability can help researchers find systematic failures of modeling. Motivated by this, in this paper we conduct representation analysis to interpret the modeling process of a U-net architecture deep learning model. We use interactive visualization to provide a comprehensive error analysis of the predictions and find underlying errors in the data. We build an online interactive panel using shiny app where users could interact with the panel to discover the prediction results.

The rest of the paper is organized as follows. In Section \ref{section:Background}, we introduce several concepts used throughout. In Section \ref{section:Preprocess}, we discuss data preprocessing. Then we introduce details of our model and model training process. Both steps are guided and explained by data visualization. In Section \ref{section:Representation}, we present the interpretation and analysis of the trained model using representation analysis. In Section \ref{section:Error}, we introduce a shiny app for error analysis and discuss the findings with the help of our app. In Section \ref{section:Conclusion}, we conclude our work and discuss how to reproduce our work for further research.

\section{Background}
\label{section:Background}
We first review background on interactive visualization and representation analysis of deep learning models.

\subsection{Interactive Visualization}
The large scale of remote sensing data -- both in the scope of area covered and number of sensor features -- creates challenges for data exploration. An increasing number of interactive visualization techniques have been proposed to support geospatial and remote sensing research.
Focusing and linking are two techniques for high-dimensional data visualization \cite{buja1991interactive}. Focusing supports visualization of only a part of the data in a single view. Focusing visualization techniques include subset selection and dimension reduction. They are commonly applied by zooming, panning, slicing, projecting and data reduction. Focusing limits the amount of presented information. In contrast, to have a more comprehensive understanding of data, linking can be used to display multiple views of data together. Different views are synchronized to give a more extensive description of the data. Linked visualizations are often implemented by brushing, clicking and dragging. 

For example, \citet{anselin2002visualizing} dynamically link a cartographic representation of data on a map with summary statistical graphics, like histograms, box plots, and scatterplots. Their interface implements linking and brushing between maps and statistical graphics. \citet{anselin1996interactive} presents an interactive dynamic framework in which brushing across a variogram cloud plot highlights pairs of observations on the map, suggesting potential spatial outliers. \citet{wisc_visualize} implements rotation, zooming, panning in three dimensions. It also allows users to select combinations of scalar variables and user could interactively control time animation of data. In \cite{tasnim2020data}, a data reduction method is proposed to reduce data size by 75\% while preserving the visual elements of images. 

\subsection{U-Net Model}

The U-Net deep learning architecture has been widely used in image segmentation problem \cite{ronneberger2015u}. It is also widely used in glacier segmentation \cite{he2020glacier,baraka2020machine,holzmann2021glacier}. U-Net model contain two parts: an encoder and a decoder. The encoder contains down-sampling layers and the decoder contains up-sampling layers. Features are extracted by the down-sampling layers at pixel levels. The extracted information is then up-sampled to the input resolution using the decoder. Skip connections between down-sampling layers and corresponding up-sampling layers provide an alternative way of learning features. Through them, features can bypass downsampling and be used directly for prediction. However, the semantic information is normally learnt by deeper layers. 

Activations are formed at each layer and are used in representation analysis. They are computed as nonlinear transformations of outputs from convolution layers. From the activations, we can discover how the input is changed and gain information about which features are learnt in each U-Net layer.

\subsection{Representation Analysis}
Though they often achieve state-of-the-art performance on earth observation problems, deep neural networks are in a way black boxes since their decision rules are not easily described. Understanding deep models can inspire improvements on data analysis problems as well as methodology \cite{yosinski2015understanding}. To describe what a trained deep neural network model has learned, some researchers study the parameters of neurons at each layer \cite{mahendran2015understanding,erhan2010does}. For example, \citet{luo2016understanding} visualizes weights of all units in a deep model for digit classification. The outcome look like strokes, with clearer borders as the model goes deeper. Others aim to interpret the functions of neurons \cite{olah2017feature}. For example, one approach is to investigate what certain units are looking for by generating artificial inputs that maximizes an individual neuron's activation \cite{erhan2009visualizing}. Alternatively, it is possible to study the activations of each neuron after passing certain data through the model, whose results can reflect on the input data and allows for further unsupervised investigation \cite{simonyan2013deep,raghu2017svcca}.

\section{Data Preprocessing and Modeling}

\label{section:Preprocess}
In this section, we introduce our data preprocessing and modeling procedures. We focus on using visualization techniques to guide decisions and provide sanity check for results.
\subsection{Data Preprocessing}
\begin{figure}[H]
\includegraphics[width=\textwidth]{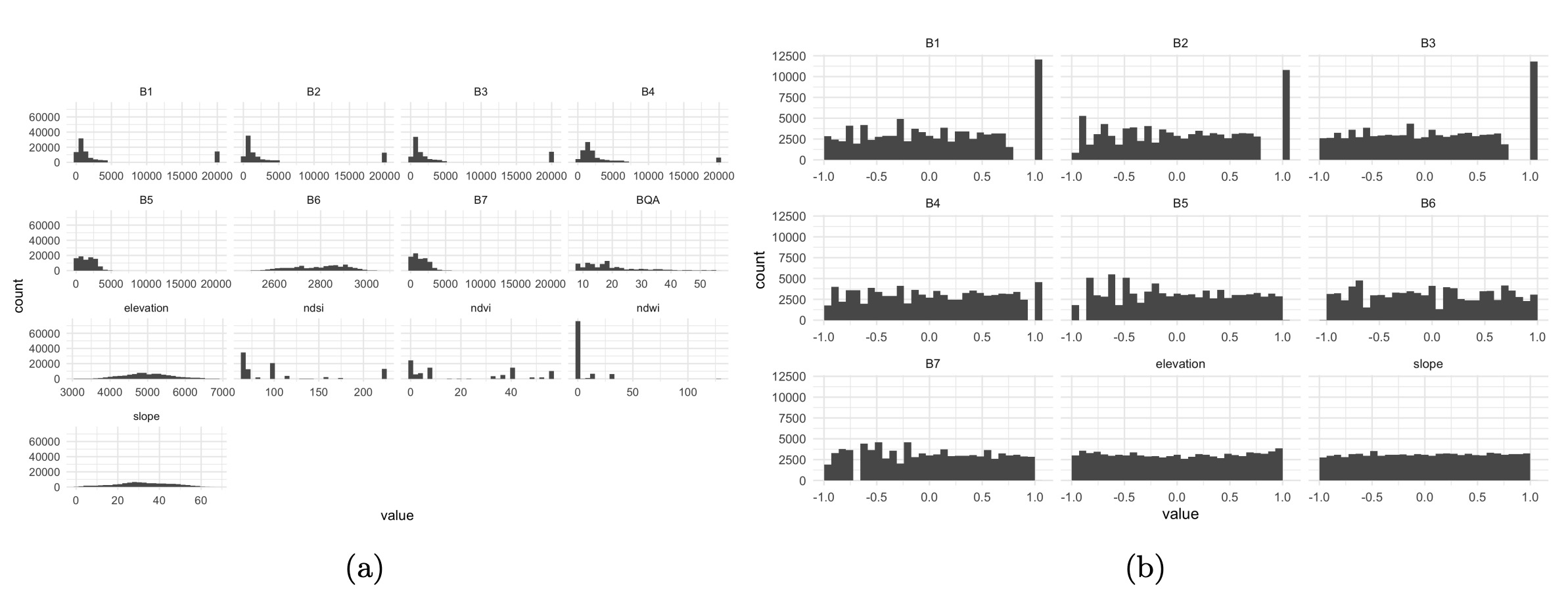}
\caption{(a) Feature histogram before preprocess. (b) Feature histogram after preprocess. Each feature before preprocessed has significantly different ranges. Each feature is equalized into the same range from -1 to 1.\label{fig:preprocess}}
\end{figure}

Our raw data contains 13 bands from Landsat 7 (LE7) and the Shuttle Radar Topography Mission (SRTM). It contains bands from B1 to B7, BQA of LE7, elevation from SRTM, and derived Normalized Difference Snow Index (NDSI), Normalized Difference Vegetation Index (NDVI), Normalized Difference Water Index (NDWI), and slope. Bands B1 to B3 are RGB color bands, B4 to B7 are infrared related bands, BQA is a quality assurance bitmask band.
Figure~\ref{fig:preprocess} provides a histogram for each channel in the raw data. We find that different features have significantly different scales -- this could lead to difficulty in model training and generalization. Specifically, large weights will be needed features with small scales; otherwise they may be overlooked during training. Hence, we equalize each feature to the -1 to 1 scale. Further, we drop the BQA, NDSI, NDVI, NDWI bands since they contain little information about the glaciers. The scaled features' histograms are also given in Figure~\ref{fig:preprocess}. We can see that all features have the same scales and almost have uniform distributions. However, for B1 to B3, a bar in each histogram that stands out. This reflects the outlying bar in the raw data histogram. But comparing with the raw data, the deviate value in scaled histogram is much closer with the other values. Though simple, these histograms of raw and transformed data give reassurance that the original data are appropriately processed for model training.

\begin{figure}[H]
\includegraphics[width=\textwidth]{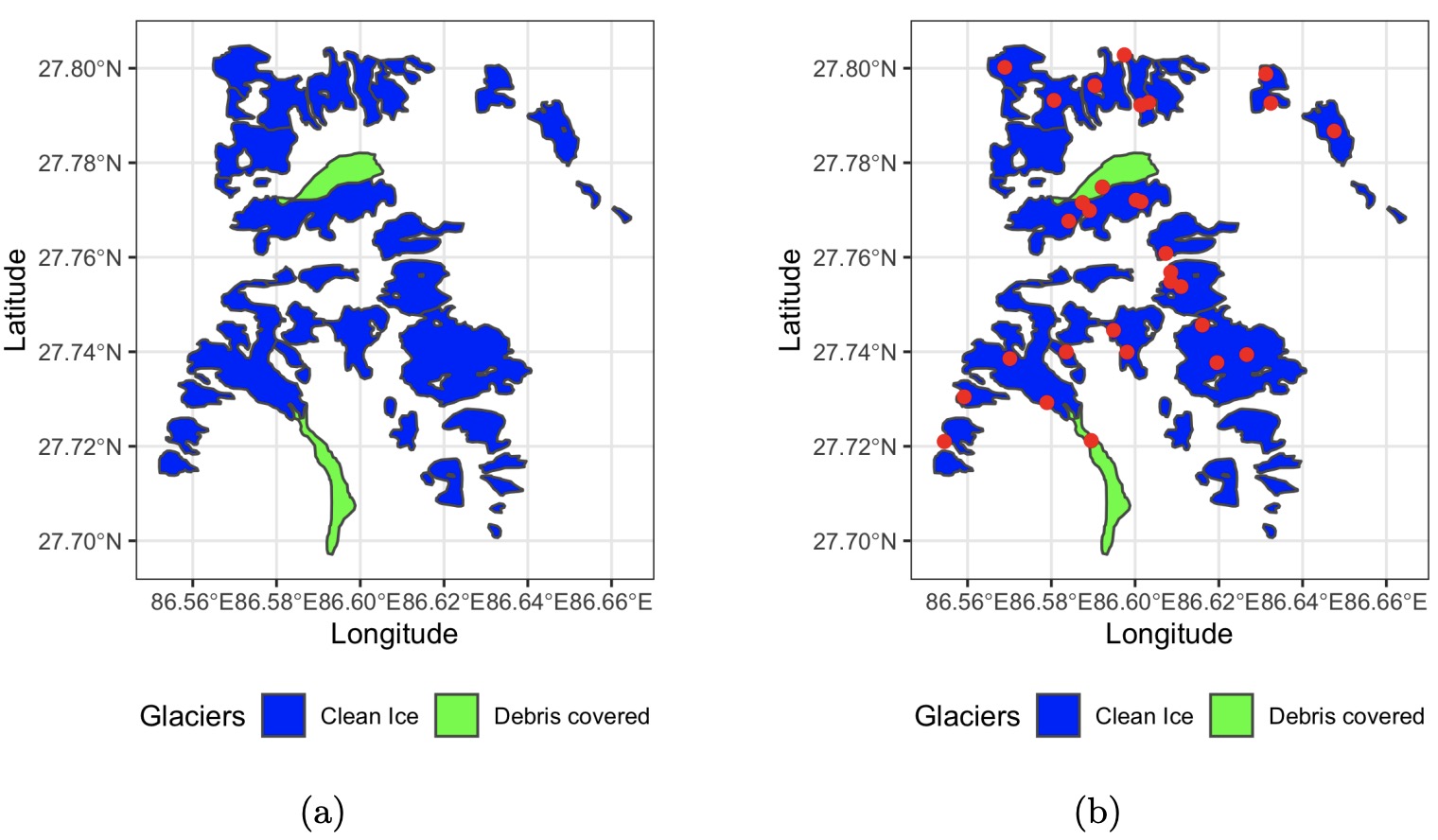}
\caption{(a) Raw glacier boundaries. (b) Glacier boundaries with sampled patches. In raw data, target glaciers only make up a limited fraction of area in the whole data. We sample patches from areas with glaciers and the center of the sampled patches are marked as red points.\label{fig:sample_glacier}}

\end{figure}

\begin{figure}[H]
\includegraphics[width=\textwidth]{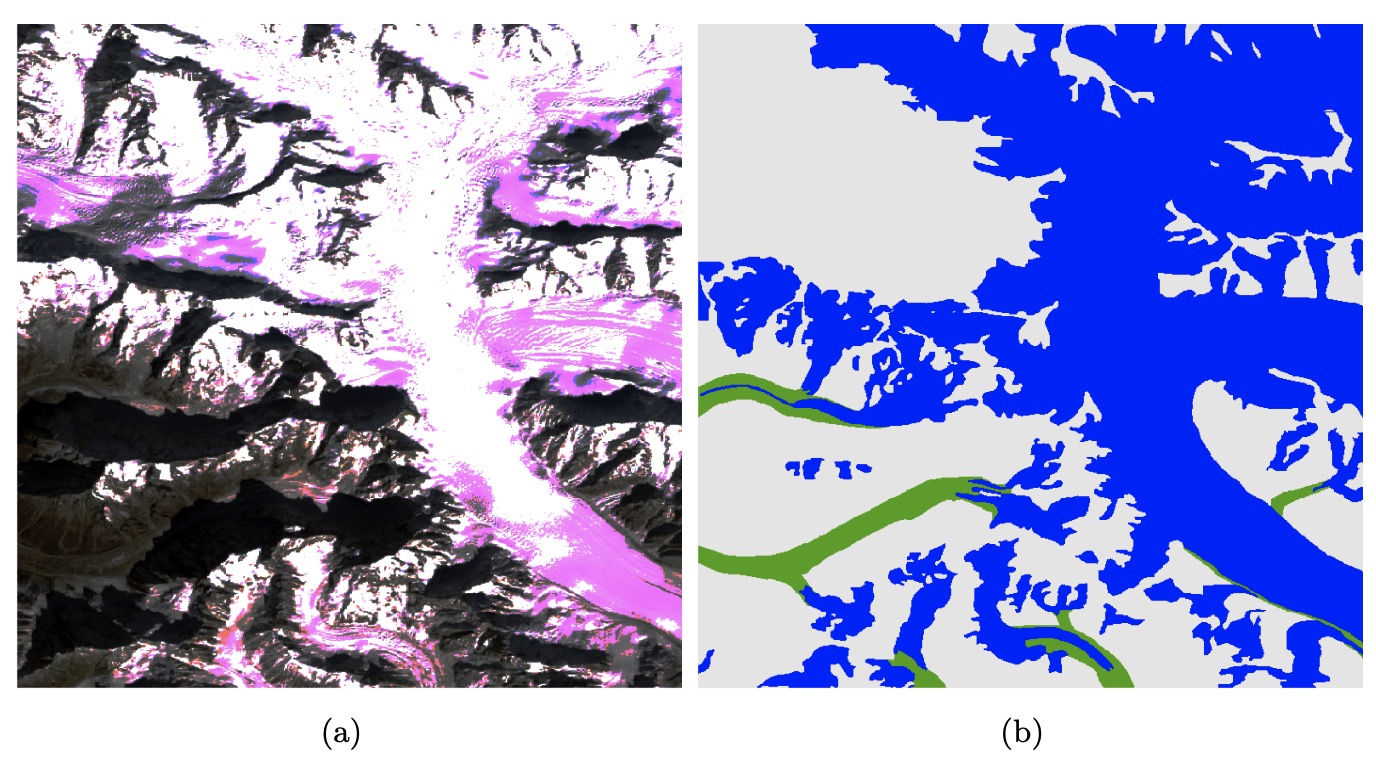}
\caption{(a) An example of preprocessed patch (b) Corresponding mask patch. In the preprocessed patch, target glaciers now make up a large proportion.\label{fig:preprocess_example}}
\end{figure}

Next, to understand the structure of the labels, we draw glacier boundaries in a partial area in Figure \ref{fig:sample_glacier}. The left panel of Figure~\ref{fig:sample_glacier} suggests a potential label imbalance issue. Compared to the non-glacier background, target glaciers make up only a small fraction of the whole area. To get more informative training data and train the model more efficiently, we randomly sample patch center points from within glaciers boundaries, not randomly across all available imagery. The resulting patches have a higher coverage rates of glaciers. The right panel of Figure \ref{fig:sample_glacier} presents example sampling results -- red points are centers of sampled patches. Note that all the red points are randomly distributed and centered on glaciers. We show example sampled patches in Figure~\ref{fig:preprocess_example}. The satellite image associated with a patch is given in Figure \ref{fig:preprocess_example} along with its ground truth mask. In the mask, we see that two types of glaciers make cover most area in the patch, reflecting the sampling strategy.%blue color represents clean ice glacier and green color represents debris covered glacier.

\subsection{Modeling}
\label{subsec:modeling}
We next introduce details about our model. We then discuss the modeling process and the ways that visualization supports it.

\begin{figure}[H]
\centering
\includegraphics[width=\textwidth]{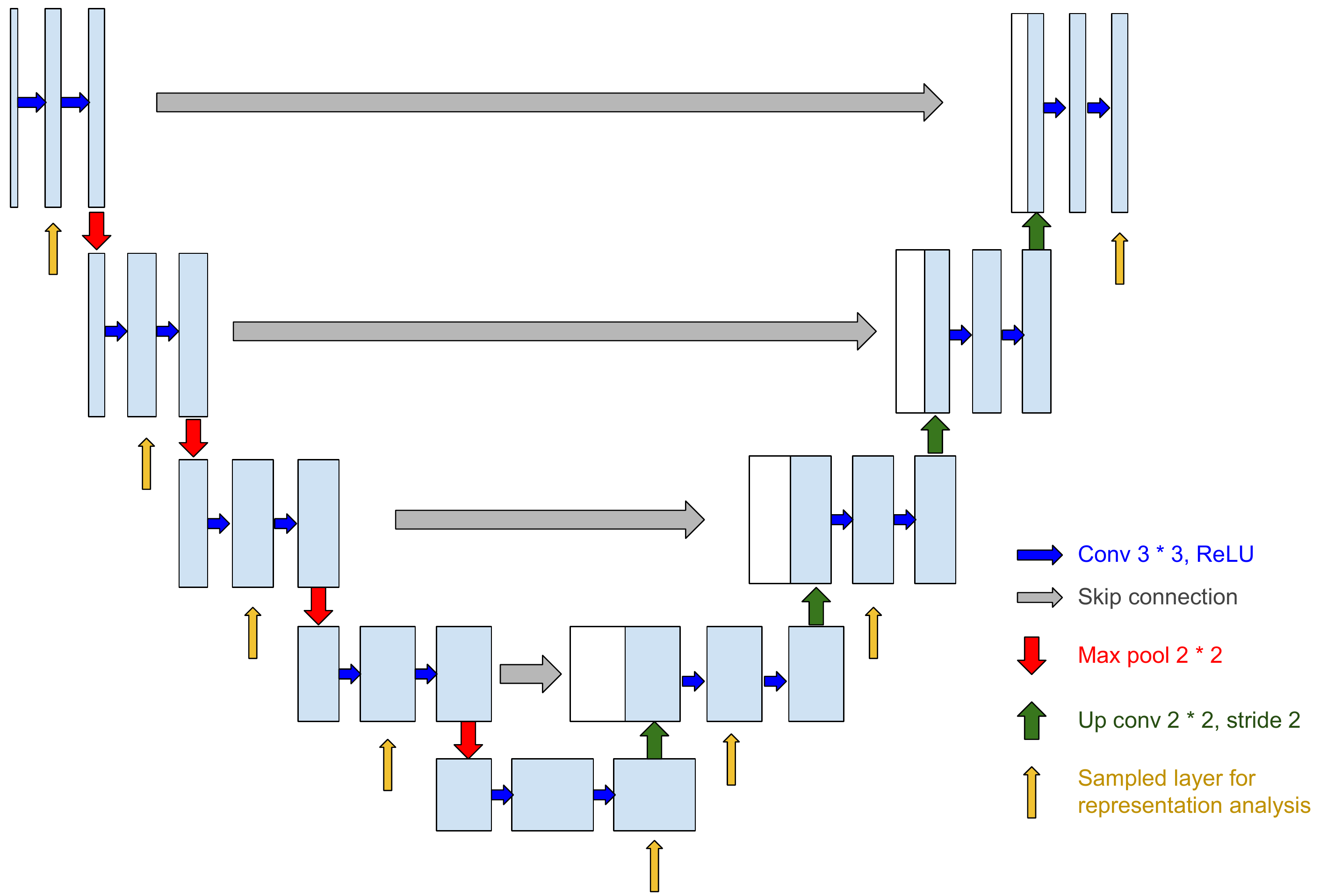}
\caption{U-Net model diagram. The arrows with different colors indicate different operations. The input is a 9-band image and output is a 3-band image with the probability of each class for each pixel.}
\label{fig:unet}
\end{figure}

We implement a U-Net architecture to the glacier segmentation task, following prior work \cite{baraka2020machine,holzmann2021glacier}. We use a kernel size of 3 $\times$ 3 with 1 padding for the convolution layers in the down-sampling blocks, middle block and up-sampling blocks. We use a kernel size of 2 $\times$ 2 with stride of 2 for the up-sampling convolution layers in the up-sampling blocks. For the pooling layer, we use maxpool with kernel size $2 \times 2$. The input data has 9 channels with size 512 $\times$ 512 and 3 output channels with the same size. The output channels correspond to clean-ice glacier, debris covered glacier, and background classes. We double the number of channels after each layer in the encoder and halve the number of channels after each layer in the decoder. The U-Net model has depth 4, with 4 down-sampling layers, 4 up-sampling layers and 1 bottleneck. During training, we use the Adam optimizer and set the learning rate to 0.0001. Dropout probability with 0.2 and $\ell_2$-regularization with $\lambda=0.0005$ are used to prevent overfitting. For the loss function, we use a combination of BCE and Dice loss. Figure \ref{fig:unet} is a diagram for our model.

\begin{figure}[H]
\centering
\includegraphics[width=10cm]{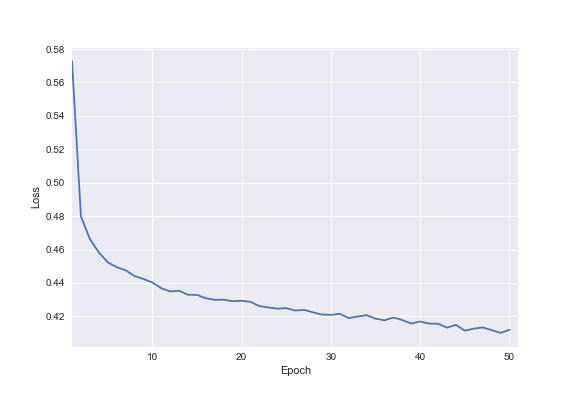}
\caption{Training loss curve. The training loss is getting smaller over epochs indicating the model is learning features from data.}
\label{fig:loss_curve}
\end{figure}

We draw the training loss curve across epochs in Figure \ref{fig:loss_curve}. In this figure, we see that from epoch 1 through 10 the loss rapidly decreases, from epoch 11 through epoch 50, the loss slowly decreases. This indicates that the model is learning properly.

\begin{figure}[H]
\includegraphics[width=\textwidth]{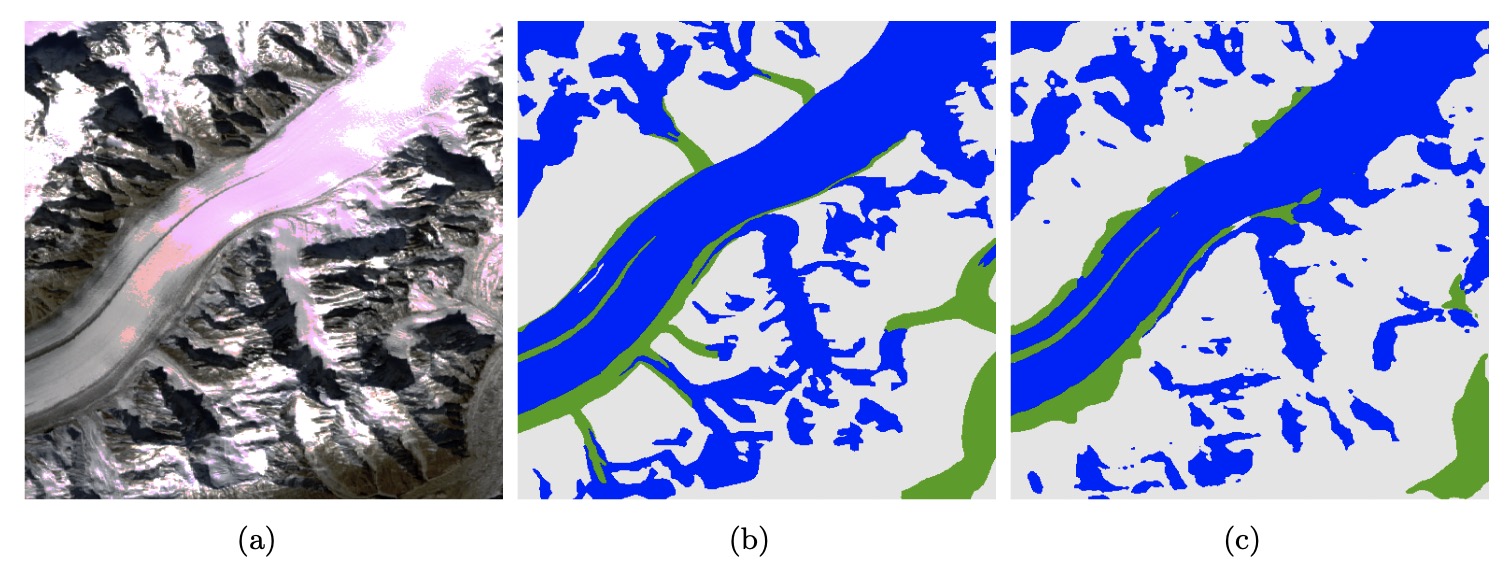}
\caption{(a) Raw image patch. (b) Mask patch. (c) Predicted patch. The model can detect the position of the bulk of target glaciers but fails to detect connections between glaciers.\label{fig:prediction_results}}
\end{figure}

Figure \ref{fig:prediction_results} provides an example model predictions. Comparing the predicted patch in the right panel with the label mask in the middle panel, it suggests that the model successfully recognize most glacier pixels in this patch. However, the model fails to accurately recognize glacier borders or to detect connections between major glacier masses. Additionally, it seems that the debris covered glaciers are not recognized well. These views provided clearer directions for improvement than an average performance metric alone.

\section{Representation Analysis}
\label{section:Representation}
We use representation analysis to understand and guide improvements for the U-Net model. In order to understand how this trained U-Net model captures underlying features and make predictions on unseen data, we investigate the functions of neurons on specific inputs by passing data through the model and visualizing the activations of each neuron.

\begin{figure}[H]
\includegraphics[width=0.9\textwidth]{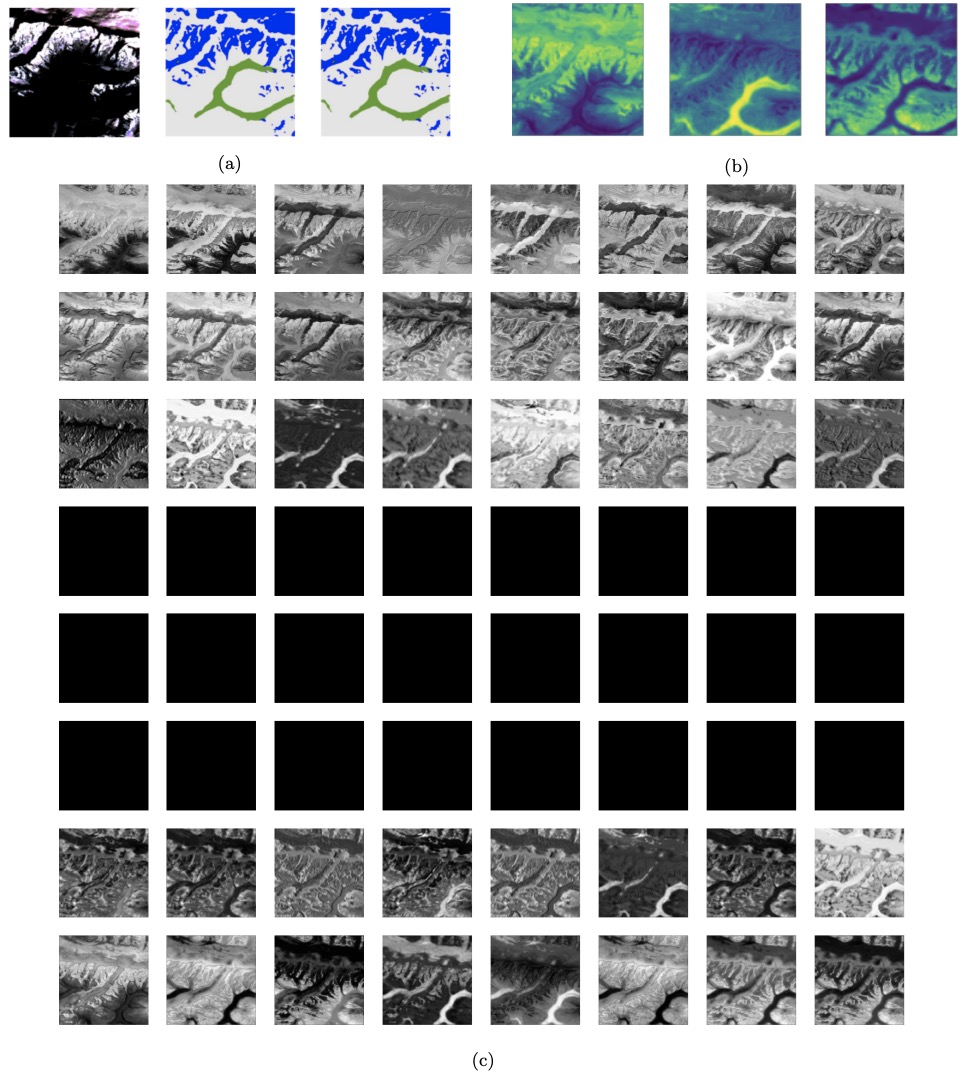}
\caption{(a) Original figure, true labels and the prediction. Blue, green, and red represents clean-ice glaciers, debris-covered glaciers, and background, respectively. (b) Visualization of the segmentation layer. The three panels represents clean ice, debris and background, respectively. (c) Activations of one satellite image across five convolutional layers of the U-Net model. Rows represent the first, third, fifth and seventh downsampling convolutional layers, the first and third upsampling convolutional layers, the last pooling layer and the second middle convolutional layer. For each layer, we randomly plot eight activations in grayscale. We observe that the activations capture basic features at the first layer, become more blurred in deeper downsampling, and become clearer at the bottleneck. Comparing activations in the same layer, we find that the activations look alike in the first layer, but different in the bottleneck.\label{fig:representation}}
\end{figure}

The U-Net model starts with down-sampling layers that condense information for distinguishing labels, followed by up-sampling layers that recover spatial context for each class. We sample activations from several layers annotated in Figure \ref{fig:unet} and present the results in Figure \ref{fig:representation}. The model-learned borders between labels become clearer and more accurate during initial down-sampling, and these activations are recovered in the last few up-sampling layers. However, the U-Net bottleneck fails to learn any information. Removing these layers may result in comparable performance with less computations. Consistent with this observation, recent literature have suggested that U-Net does not seem to learn long-range spatial relationships. For example, \cite{malkin2020mining} observes that a segmentation model trained on a landcover segmentation misclassifies roads when they are interrupted by trees.

\section{Error Analysis of Prediction Results}
\label{section:Error}
In this section, we introduce a visualization interface that supports comprehensive error analysis of prediction results.

\subsection{Interactive Panel Introduction} 

\begin{figure}[H]
\centering
\includegraphics[width=\textwidth]{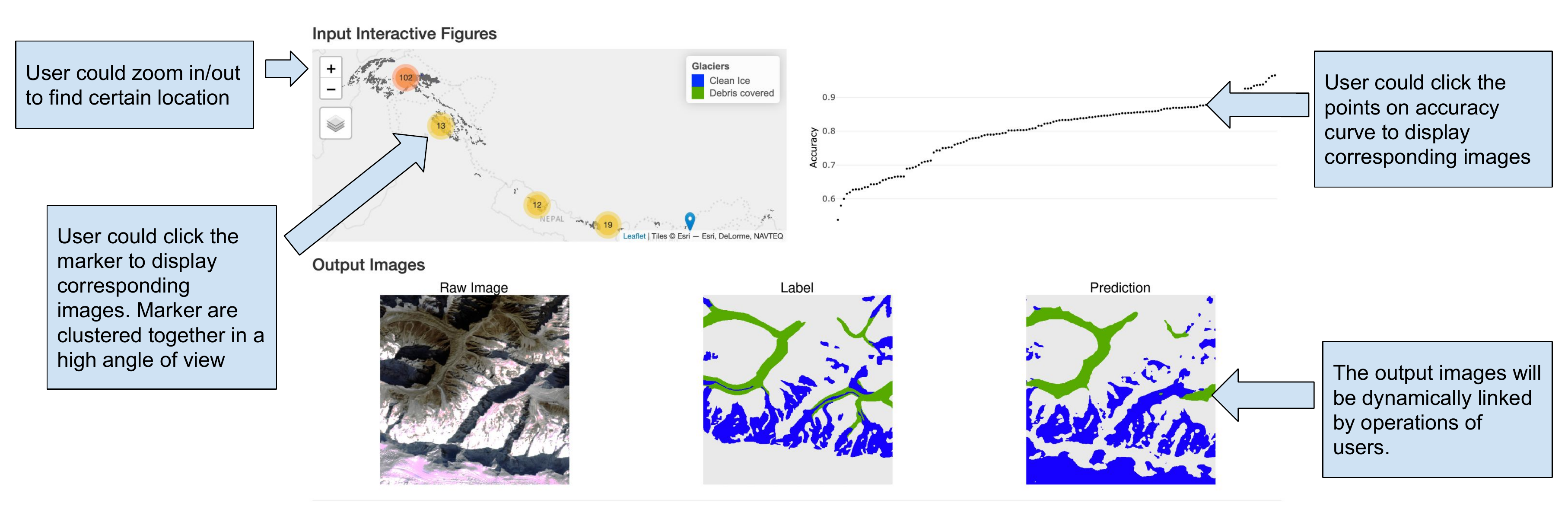}
\caption{Screenshot of shiny app with annotations. Users could interact with the app through the upper part of the app. User could click the map and loss curve to switch the prediction results in the lower part of the app.}
\label{fig:annotation_app}
\end{figure}

We build an \href{https://bruce-zheng.shinyapps.io/glacier_segmententation/}{online interactive panel} with the Shiny package \cite{shiny} in R. It allows users to interact with the glacier map and a accuracy curve. These panels are both dynamically linked to views of the corresponding area, including a raw satellite image, ground truth mask and prediction result. Figure \ref{fig:annotation_app} presents an annotated screenshot of our application. In the upper panels, users can zoom in/out to select specific region of interest. By clicking the marker in the map, users can link to the corresponding images below. The markers cluster together when the user zooms out.  We also draw an accuracy curve for test patches. Accuracy is calculated pixelwise across each patch. Clicking a point in the curve displays the corresponding patch. Intuitively, this accuracy curve lets users easily view test patches with good or poor performances, facilitating a more comprehensive understanding of model predictions. For example, from patches with low accuracy, users can identify areas where the model fails and specific glacier patterns on which the model has poor performance.

\subsection{Error Analysis Discussion}
\begin{figure}[H]
\centering
\includegraphics[width=\textwidth]{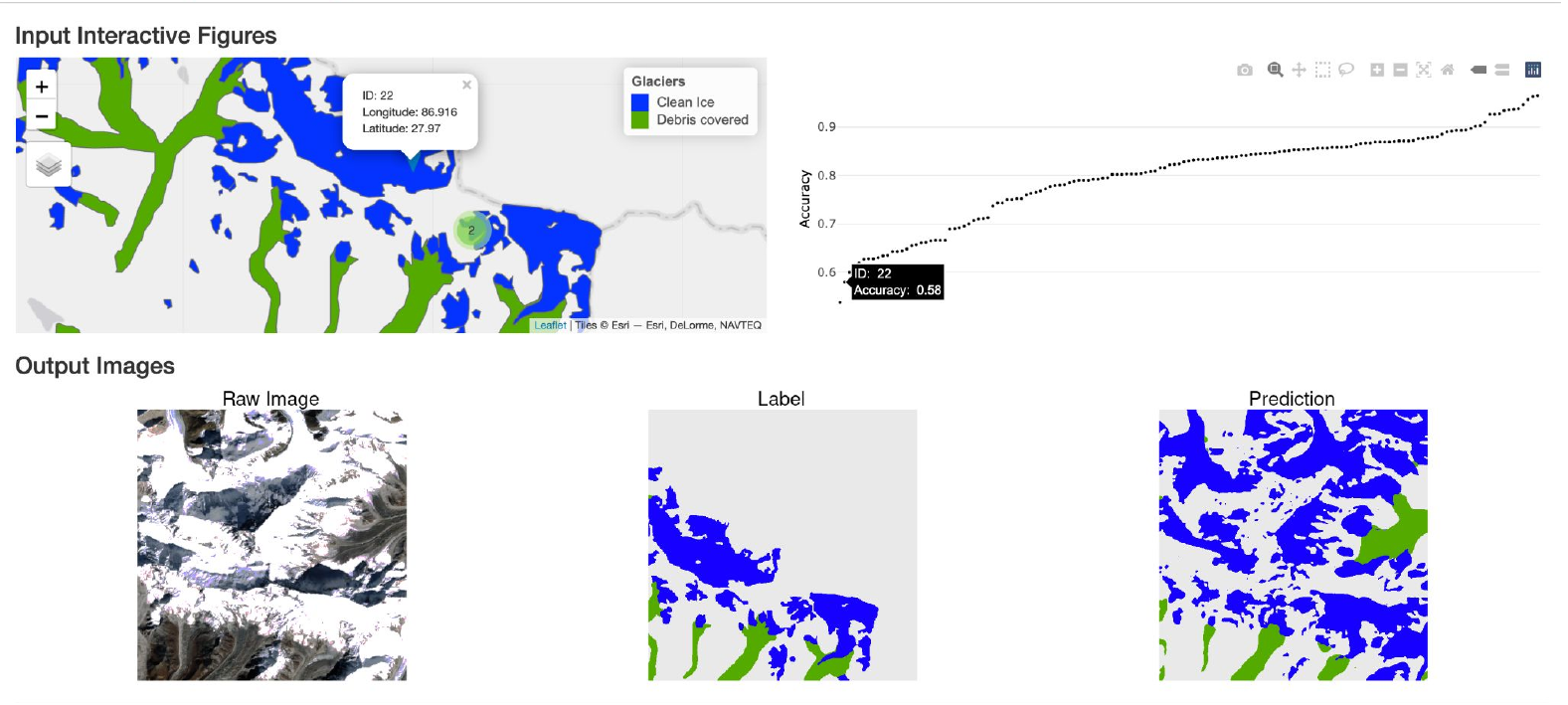}
\caption{An example of prediction results with low accuracy. We notice that the prediction result is actually similar to the raw image however the problem it that the label mask is incomplete i.e. the upper right part is missing. Comparing with the map, we find that the missing-label area is within Chinese border. The glaciers within Chinese border are not labeled.}
\label{fig:discussion}
\end{figure}

We illustrate the use of this interactive visualization to support error analysis of glacier segmentation results. Figure \ref{fig:discussion} presents an example of a test patch with a relatively low accuracy. Surprisingly, the model prediction result seems to match the raw satellite image. The reason why it has a low accuracy is that the label mask is not fully labeled -- labels are only available in Nepal, not China. Comparing with the map, we find that the upper right part of this area is within Chinese border. Since this patch includes substantial area within Chinese border, many predictions are mistakenly declared as false positives. This structure is straightforward to discover with the help of this interactive panel, but it would require a more deliberate effort without it.

\section{Conclusion}
\label{section:Conclusion}
We have used visualization to provide guidance for machine-learning supported glacier segmentation. During preprocessing, we detected potential issues in the raw data, such as label imbalance and divergent feature scales. We also provided sanity checks of the preprocessed data to demonstrate their appropriateness for modeling. In Section \ref{subsec:modeling}, the training loss curve suggests that the model is learns properly. To better understand trainign reuslts, we visualize model activations. Based on the activations, we find that the deepest layers are bypassed, with most activations flowing through skip connections. To support prediction error analysis, we built an online interactive visualization to display and critique  prediction results. Users can easily interact with the app to discover patterns across patch prediction results. We shared an example where the app revealed an issue with the source labels.

We release our code at \href{https://github.com/krisrs1128/geo_mlvis}{github repository}. We provide the code for data preprocessing, model training and inference, representation analysis and shiny app definition. We hope this code can be re-used by others seeking visualization or representation analysis of geospatial deep learning models.

\authorcontributions{Formal analysis, Minxing Zheng and Xinran Miao; Project administration, Kris Sankaran; Supervision, Kris Sankaran; Validation, Kris Sankaran; Visualization, Minxing Zheng and Xinran Miao; Writing – original draft, Minxing Zheng and Xinran Miao; Writing – review \& editing, Minxing Zheng, Xinran Miao and Kris Sankaran.}

\acknowledgments{We would like to express our thankfulness to all the members in our \href{https://krisrs1128.github.io/LSLab/}{Latent Structure Lab} for their questions during preliminary presentations and feedback on drafts. We gained much help from them and extended our ideas in many aspects of this paper. Thanks for Prof. Keith Levin and Prof. Hyunseung Kang at UW-Madison who provided great suggestions for the shiny app. We thank The Center for High Throughput Computing (CHTC) at UW-Madison, which provided our computational resources.}

%%%%%%%%%%%%%%%%%%%%%%%%%%%%%%%%%%%%%%%%%%
%% Optional
\appendixtitles{no} % Leave argument "no" if all appendix headings stay EMPTY (then no dot is printed after "Appendix A"). If the appendix sections contain a heading then change the argument to "yes".
\appendixstart
\appendix
\section[\appendixname~\thesection]{}
\begin{figure}[H]
\includegraphics[width=\textwidth]{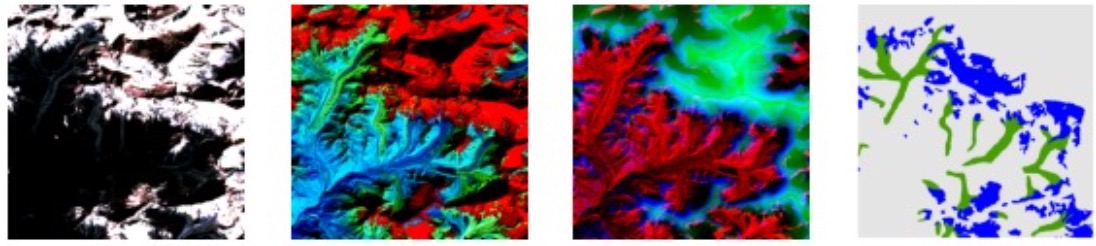}
\caption{An example of input channels and output labels. We visualize the nine input channels (first three columns) in groups of three and the output labels (last column). Blue, green and grey represents clean-ice glaciers, debris - covered glaciers and background, respectively.\label{fig:appendix}}
\end{figure}

%\subsection[\appendixname~\thesubsection]{}
Figure \ref{fig:appendix} gives an example of the input patches and output masks from raw data. The nine channels complement information with each other in the sense of distinguishable borders between labels as well as possible missing information in some channels.

%%%%%%%%%%%%%%%%%%%%%%%%%%%%%%%%%%%%%%%%%%
\begin{adjustwidth}{-\extralength}{0cm}

\reftitle{References}

\bibliography{ref.bib}

\begin{thebibliography}{999}

\bibitem[Buja \em{et~al.}(1991)Buja, McDonald, Michalak, and
  Stuetzle]{buja1991interactive}
Buja, A.; McDonald, J.A.; Michalak, J.; Stuetzle, W.
\newblock Interactive data visualization using focusing and linking.
\newblock  Proceedings of the 2nd conference on Visualization'91,  1991, pp.
  156--163.

\bibitem[Anselin \em{et~al.}(2002)Anselin, Syabri, Smirnov,
  et~al.]{anselin2002visualizing}
Anselin, L.; Syabri, I.; Smirnov, O.;  et~al.
\newblock Visualizing multivariate spatial correlation with dynamically linked
  windows.
\newblock  Proceedings, CSISS Workshop on New Tools for Spatial Data Analysis,
  Santa Barbara, CA. Citeseer,  2002.

\bibitem[Anselin(1996)]{anselin1996interactive}
Anselin, L.
\newblock Interactive techniques and exploratory spatial data analysis {\bf
  1996}.

\bibitem[Hibbard and Santek(1989)]{wisc_visualize}
Hibbard, W.; Santek, D.
\newblock Visualizing large data sets in the earth sciences.
\newblock {\em Computer} {\bf 1989}, {\em 22},~53--57.
\newblock
  doi:{\changeurlcolor{black}\href{https://doi.org/10.1109/2.35200}{\detokenize{10.1109/2.35200}}}.

\bibitem[Tasnim and Mondal(2020)]{tasnim2020data}
Tasnim, J.; Mondal, D.
\newblock Data Reduction and Deep-Learning Based Recovery for Geospatial
  Visualization and Satellite Imagery.
\newblock  2020 IEEE International Conference on Big Data (Big Data). IEEE,
  2020, pp. 5276--5285.

\bibitem[Ronneberger \em{et~al.}(2015)Ronneberger, Fischer, and
  Brox]{ronneberger2015u}
Ronneberger, O.; Fischer, P.; Brox, T.
\newblock U-net: Convolutional networks for biomedical image segmentation.
\newblock  International Conference on Medical image computing and
  computer-assisted intervention. Springer,  2015, pp. 234--241.

\bibitem[He \em{et~al.}(2020)He, Zhang, Ma, and Wu]{he2020glacier}
He, Q.; Zhang, Z.; Ma, G.; Wu, J.
\newblock GLACIER IDENTIFICATION FROM LANDSAT8 OLI IMAGERY USING DEEP U-NET.
\newblock {\em ISPRS Annals of Photogrammetry, Remote Sensing \& Spatial
  Information Sciences} {\bf 2020}, {\em 5}.

\bibitem[Baraka \em{et~al.}(2020)Baraka, Akera, Aryal, Sherpa, Shresta, Ortiz,
  Sankaran, Ferres, Matin, and Bengio]{baraka2020machine}
Baraka, S.; Akera, B.; Aryal, B.; Sherpa, T.; Shresta, F.; Ortiz, A.; Sankaran,
  K.; Ferres, J.L.; Matin, M.; Bengio, Y.
\newblock Machine Learning for Glacier Monitoring in the Hindu Kush Himalaya,
  2020,  \href{http://xxx.lanl.gov/abs/2012.05013}{{\normalfont
  [arXiv:cs.CV/2012.05013]}}.

\bibitem[Holzmann \em{et~al.}(2021)Holzmann, Davari, Seehaus, Braun, Maier, and
  Christlein]{holzmann2021glacier}
Holzmann, M.; Davari, A.; Seehaus, T.; Braun, M.; Maier, A.; Christlein, V.
\newblock Glacier Calving Front Segmentation Using Attention U-Net,  2021,
  \href{http://xxx.lanl.gov/abs/2101.03247}{{\normalfont
  [arXiv:cs.LG/2101.03247]}}.

\bibitem[Yosinski \em{et~al.}(2015)Yosinski, Clune, Nguyen, Fuchs, and
  Lipson]{yosinski2015understanding}
Yosinski, J.; Clune, J.; Nguyen, A.; Fuchs, T.; Lipson, H.
\newblock Understanding neural networks through deep visualization.
\newblock {\em arXiv preprint arXiv:1506.06579} {\bf 2015}.

\bibitem[Mahendran and Vedaldi(2015)]{mahendran2015understanding}
Mahendran, A.; Vedaldi, A.
\newblock Understanding deep image representations by inverting them.
\newblock  Proceedings of the IEEE conference on computer vision and pattern
  recognition,  2015, pp. 5188--5196.

\bibitem[Erhan \em{et~al.}(2010)Erhan, Courville, Bengio, and
  Vincent]{erhan2010does}
Erhan, D.; Courville, A.; Bengio, Y.; Vincent, P.
\newblock Why does unsupervised pre-training help deep learning?
\newblock  Proceedings of the thirteenth international conference on artificial
  intelligence and statistics. JMLR Workshop and Conference Proceedings,  2010,
  pp. 201--208.

\bibitem[Luo \em{et~al.}(2016)Luo, Li, Urtasun, and
  Zemel]{luo2016understanding}
Luo, W.; Li, Y.; Urtasun, R.; Zemel, R.
\newblock Understanding the effective receptive field in deep convolutional
  neural networks.
\newblock  Proceedings of the 30th International Conference on Neural
  Information Processing Systems,  2016, pp. 4905--4913.

\bibitem[Olah \em{et~al.}(2017)Olah, Mordvintsev, and
  Schubert]{olah2017feature}
Olah, C.; Mordvintsev, A.; Schubert, L.
\newblock Feature visualization.
\newblock {\em Distill} {\bf 2017}, {\em 2},~e7.

\bibitem[Erhan \em{et~al.}(2009)Erhan, Bengio, Courville, and
  Vincent]{erhan2009visualizing}
Erhan, D.; Bengio, Y.; Courville, A.; Vincent, P.
\newblock Visualizing higher-layer features of a deep network.
\newblock {\em University of Montreal} {\bf 2009}, {\em 1341},~1.

\bibitem[Simonyan \em{et~al.}(2013)Simonyan, Vedaldi, and
  Zisserman]{simonyan2013deep}
Simonyan, K.; Vedaldi, A.; Zisserman, A.
\newblock Deep inside convolutional networks: Visualising image classification
  models and saliency maps.
\newblock {\em arXiv preprint arXiv:1312.6034} {\bf 2013}.

\bibitem[Raghu \em{et~al.}(2017)Raghu, Gilmer, Yosinski, and
  Sohl-Dickstein]{raghu2017svcca}
Raghu, M.; Gilmer, J.; Yosinski, J.; Sohl-Dickstein, J.
\newblock Svcca: Singular vector canonical correlation analysis for deep
  learning dynamics and interpretability.
\newblock {\em arXiv preprint arXiv:1706.05806} {\bf 2017}.

\bibitem[Malkin \em{et~al.}(2020)Malkin, Ortiz, and Jojic]{malkin2020mining}
Malkin, N.; Ortiz, A.; Jojic, N.
\newblock Mining self-similarity: Label super-resolution with epitomic
  representations.
\newblock  European Conference on Computer Vision. Springer,  2020, pp.
  531--547.

\bibitem[Chang \em{et~al.}(2021)Chang, Cheng, Allaire, Sievert, Schloerke, Xie,
  Allen, McPherson, Dipert, and Borges]{shiny}
Chang, W.; Cheng, J.; Allaire, J.; Sievert, C.; Schloerke, B.; Xie, Y.; Allen,
  J.; McPherson, J.; Dipert, A.; Borges, B.
\newblock {\em shiny: Web Application Framework for R},  2021.
\newblock R package version 1.7.1.

\end{thebibliography}

\end{adjustwidth}
\end{document}